\documentclass[10pt,twocolumn,letterpaper]{article}

\usepackage{iccv}
\usepackage{times}
\usepackage{epsfig}
\usepackage{graphicx}
\usepackage{amsmath}
\usepackage{amssymb}

\usepackage{algorithmic}
\usepackage{algorithm}
\usepackage{multirow}
\usepackage{subfigure}
\usepackage{authblk}

\usepackage[breaklinks=true,bookmarks=false]{hyperref}

\usepackage{lipsum}

\newcommand{\chapternote}[1]{{%
  \let\thempfn\relax 
  \footnotetext[0]{#1} 
}}

\iccvfinalcopy 

\def\iccvPaperID{305} 

\ificcvfinal\fi
\begin{document}

\title{Joint Layout Estimation and Global Multi-View Registration \\ for Indoor Reconstruction}

\author[1]{Jeong-Kyun Lee\thanks{Denote equal contribution.}}
\newcommand\CoAuthorMark{\footnotemark[\arabic{footnote}]} 
\author[2]{Jaewon Yea\protect\CoAuthorMark}
\author[3]{Min-Gyu Park}
\author[1]{Kuk-Jin Yoon\thanks{Corresponding author.}}
\affil[1]{Computer Vision Laboratory, GIST, Gwangju, South Korea}
\affil[2]{LG Electronics, Incheon, South Korea}
\affil[3]{Korea Electronics Technology Institute, Seongnam, South Korea \protect\\	{\tt\small \{leejk,kjyoon\}@gist.ac.kr, jaewon.yea@lge.com, mpark@keti.re.kr}}

\maketitle

\begin{abstract}
In this paper, we propose a novel method to jointly solve scene layout estimation and global registration problems for accurate indoor 3D reconstruction. Given a sequence of range data, we first build a set of scene fragments using KinectFusion and register them through pose graph optimization. Afterwards, we alternate between layout estimation and layout-based global registration processes in iterative fashion to complement each other. We extract the scene layout through hierarchical agglomerative clustering and energy-based multi-model fitting in consideration of noisy measurements. Having the estimated scene layout in one hand, we register all the range data through the global iterative closest point algorithm where the positions of 3D points that belong to the layout such as walls and a ceiling are constrained to be close to the layout. We experimentally verify the proposed method with the publicly available synthetic and real-world datasets in both quantitative and qualitative ways. 
\end{abstract} \vspace{-10pt}

\section{Introduction} \chapternote{This work was done when Jaewon Yea and Min-Gyu Park were members of the Computer Vision Laboratory at GIST.}

The  popularization of low-cost consumer depth cameras has made a new perspective of solving various computer vision problems. Especially, with various depth sensors, simultaneous localization and mapping (SLAM) and 3D reconstruction~\cite{Choi:2015, Newcombe:2011} have shown visually compelling results compared to conventional image-based approaches in an indoor environment. This is because the consumer depth camera robustly acquires depth measurements where the conventional image-based approaches frequently fail to estimate accurate depth, \eg, due to poorly textured regions. In this paper, we narrow our attention to the complete 3D reconstruction problem in an indoor environment using range measurements acquired from a consumer depth camera.

\begin{figure}[t]
	\centering \hspace{-1.0mm}
	\includegraphics[width=0.95\columnwidth,height=190pt]{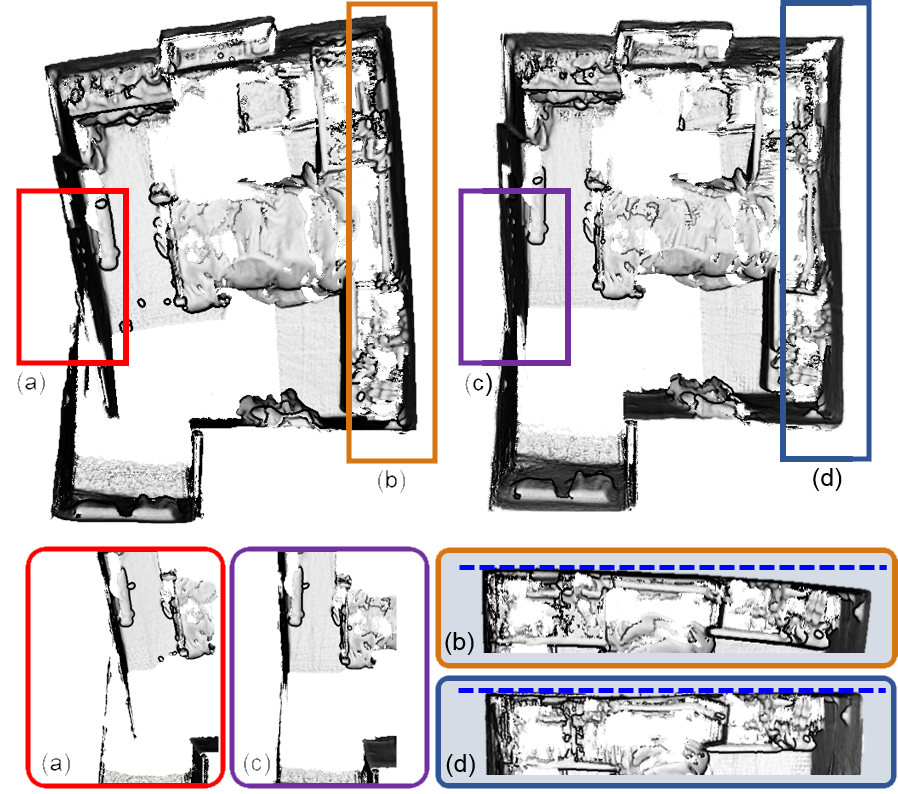}
	\caption{Comparison of the proposed method (right) with the state-of-the-art method~\cite{Choi:2015} (left). With the aid of the scene layout, the proposed method preserves important structures of the scene such as walls and a floor.}\label{fig:clustered}\vspace{-17.0pt}
\end{figure} 


KinectFusion~\cite{Newcombe:2011}, one of the pioneering works, showed that a real-world object as well as an indoor scene can be reconstructed in real-time with GPU acceleration. It exploits the iterative closest point (ICP) algorithm~\cite{Besl:1992} to track 6-DoF poses and the volumetric surface representation scheme with signed distance functions~\cite{Curless:1996} to fuse 3D measurements. A number of following studies~\cite{Choi:2015, Whelan:2012, Whelan:2015} have tackled the limitation of KinectFusion; as the scale of a scene increases, it is hard to completely reconstruct the scene due to the drift problem of the ICP algorithm as well as the large memory consumption of volumetric integration.
To scale up the KinectFusion algorithm, Whelan \etal~\cite{Whelan:2012} presented a spatially extended KinectFusion, named as Kintinuous, by incrementally adding KinectFusion results as the form of triangular meshes. Moreover, they used a pose graph to alleviate the drift problem through graph optimization by identifying loop closures. Whelan \etal~\cite{Whelan:2015} also proposed ElasticFusion
to overcome the problem 
by using the surface loop closure optimization and the surfel-based representation.
On the other hand, this large-scale indoor reconstruction problem has been tackled from the view point of global registration~\cite{Arrigoni:2016, Choi:2015, Shiratori:2015}. Notably, Choi \etal~\cite{Choi:2015} showed promising results. They utilized KinectFusion results as building blocks and developed a robust global registration scheme based on line-processes in the presence of sensor noise. 

Furthermore, the large-scale indoor reconstruction problem has been tackled by considering the structural regularities of an indoor scene such as the axis-aligned geometry~\cite{Xiao:2012} and the planarity of the scene~\cite{Ma:2016, Moreno:2014, Zhang:2015}. Xiao and Furukawa~\cite{Xiao:2012} showed that a museum-level indoor environment can be effectively reconstructed based on the Manhattan world assumption, \eg walls, floors, and ceilings are parallel to one of the three orthogonal surfaces. However, this strong assumption about the scene generates oversimplified structures in practical situations. To avoid the potential oversimplification problem, Zhang \etal~\cite{Zhang:2015} analyzed planar and non-planar regions on the fly and integrated KinectFusion's results seamlessly.  

In this paper, we pose a new approach for accurate indoor reconstruction by jointly resolving the scene layout estimation problem and the global registration problem of range data. Given initially registered range data, we extract the envelope of an indoor scene, including walls, a floor, and a ceiling, through hierarchical agglomerative clustering and energy-based multi-model fitting to reduce redundant planes and to find dominant plane hypotheses. Then, we establish point-to-layout correspondences to constrain the position of these correspondences to be close to the layout. Finally, we register entire range data through the global ICP algorithm with pairwise and layout-based constraints. We repeat layout estimation and layout-based global registration procedures alternately until they converge. Note that we purely rely on range data and the layout is computed with a weak Manhattan world assumption, such that walls are not necessarily perpendicular to each other but perpendicular to the floor and the ceiling.

\section{Previous Work}

The literature review primarily focuses on indoor 3D reconstruction starting from KinectFusion~\cite{Newcombe:2011} and its follow-up researches that aim at scaling-up KinectFusion. 

After the breakthrough of Newcombe \etal~\cite{Newcombe:2011}, hundreds of papers have addressed the limitations of KinectFusion. To overcome the scalability of the volumetric reconstruction approach, some works~\cite{Whelan:2012, Roth:2012, Heredia:2012} adopted the concept of moving volume, which translates and rotates a reconstructed volume by using the estimated pose information. In the similar manner, Steinbrucker \etal~\cite{Steinbrucker:2013} proposed a multi-scale octree data structure to modify the uniform volumetric structure into non-uniform volume. Henry \etal~\cite{Henry:2013} proposed a multiple-volume representation to create a globally consistent indoor environment. Chen \etal~\cite{Chen:2013} and Nie{\ss}ner \etal~\cite{Nie:2013} proposed a memory-efficient hierarchical data structure for commodity graphics hardware to extend KinectFusion to large-scale scenes. 

On the other hand, numerous researches focused on alleviating accumulation errors in reconstructing a large-scale environment. Conventional pairwise approaches~\cite{Chen:1992, Pulli:1999} as well as KinectFusion incrementally integrate a set of range data using the ICP algorithm; they suffer from accumulated errors in general. Therefore, the global registration approaches~\cite{Bergevin:1996, Williams:2000, Nishino:2002, Borrmann:2008, Shiratori:2015, Gelfand:2005,Liu:2014,Tang:2015,Arrigoni:2016} have been developed to alleviate accumulation errors by optimizing the global poses simultaneously. Bergevin \etal~\cite{Bergevin:1996} proposed a star-shaped network for global registration. Each range view can be interconnected with the world reference frame by sequential transformation multiplications. The transformations between the range views and the world frame are alternately optimized by the point-to-plane method~\cite{Chen:1992}. Nishino and Ikeuchi~\cite{Nishino:2002} proposed a robust global registration method based on the M-estimator~\cite{Huber:2011} to improve robustness against outlier correspondences. Arrigoni \etal~\cite{Arrigoni:2016} proposed the global registration method based on the low-rank and sparse (LRS) decomposition. 

For the sake of robust global registration, a number of researches~\cite{Borrmann:2008, Shiratori:2015} focused on identifying loop closures which must be acquired for global registration. As long as loop closures are properly identified, it is possible to reduce accumulation errors effectively. Several researches~\cite{Whelan:2012, Whelan:2013, Steinbrucker:2013, Henry:2013} used visual features to identify loop closures, but they showed failure cases under pose variations as well as when revisiting poorly textured regions. Therefore, some researches~\cite{Rusu:2009, Fern:2013, Cupec:2015, Choi:2015} focused  on solving this problem with geometric features. Assuming loop closures are given from a set of accurate correspondences, the pose graph optimization scheme has been widely employed~\cite{Whelan:2012, Whelan:2013, Steinbrucker:2013, Henry:2013, Choi:2015} owning to its real-time performance. It optimizes a pose graph constrained by pairwise transformations to balance the accumulated error. Tang and Feng~\cite{Tang:2015} proposed the method to distribute the accumulated error by integrating loops incrementally. They minimized the bi-directional registration errors~\cite{Liu:2014} of the virtual point pairs~\cite{Pulli:1999} using the global optimization technique~\cite{Williams:2000}.   

\begin{figure*}[t]
	\centering	\includegraphics[width=1.75\columnwidth,height=166pt]{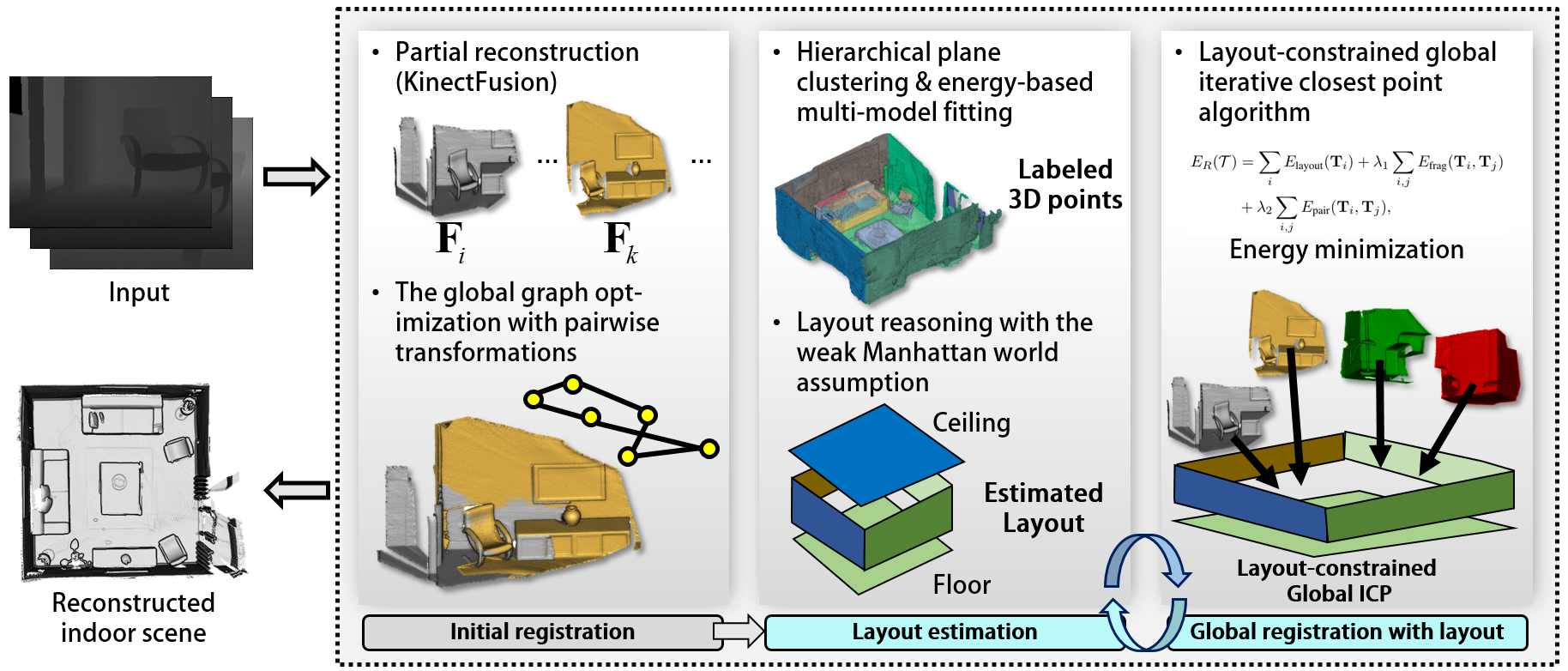}
	\caption{The overall procedure of the proposed method.}\label{fig:procedure}\vspace{-13.0pt}
\end{figure*} 
%

In addition, a number of researches focused on the structural regularity of the indoor scene to elevate the quality of reconstructed models. Basically, the planarity assumption~\cite{Dou:2012, Ataer:2013, Taguchi:2013,Zhang:2015,Moreno:2014} is the most commonly used one. Moreno \etal~\cite{Moreno:2014} proposed an incremental plane mapping scheme in which the relation between planes is identified by point features. Several studies~\cite{Dou:2012, Ataer:2013, Taguchi:2013} exploited planes and points to find frame-to-frame camera pose and to define an objective function for bundle adjustment. Ma \etal~\cite{Ma:2016} estimated a global plane model and frame-to-frame pose in an alternative way in the EM framework. Zhang \etal~\cite{Zhang:2015} proposed an interactive reconstruction algorithm, in which the algorithm guides the person to capture designated spots. 

The proposed method overcomes aforementioned problems through the layout-constrained global registration. The scene layout estimation problem has been tackled in the field of scene understanding~\cite{Choi:2013,Furlan:2013,Lee:2010} and object detection~\cite{Geiger:2011}.
Some researches~\cite{Oesau:2016,Verdie:2015,Zhou:2012} proposed to enforce the global regularity (\eg parallelism, orthogonality, and coplanarity) of the scene structures in an iterative fashion, assuming that \textit{well-aligned} but noisy point clouds are given as input. However, we consider \textit{inaccurately aligned} point clouds, \ie, owing to drift errors as shown in the left of Fig.~\ref{fig:clustered}. Therefore, we perform the layout estimation and global registration jointly, and in particular, the proposed dominant plane estimation based on energy minimization provides locally optimal dominant planes without regard to the general global regularities.


\section{Proposed Method}

We propose a joint approach of scene layout estimation and global multi-view registration for 3D indoor reconstruction. As shown in Fig.~\ref{fig:procedure}, the overall procedure of the proposed method consists of three main steps: initial registration, layout estimation, and global registration. In the initial registration step, we sequentially construct scene fragments from range data and then align them in the world coordinate system. Afterward, we alternate between layout estimation and global registration procedures in iterative fashion using scene fragments. The model reconstructed by the global registration is refined using \cite{Zhou:2014}. 

\subsection{Initial Registration} \label{sec:initreg}

For initial registration, we partially reconstruct the captured indoor scene to produce a set of scene fragments and then register them in the world coordinate system, which is similar to the previous study of Choi \etal~\cite{Choi:2015}. Here, the underlying assumption is that each scene fragment contains a negligible amount of accumulation errors so that the large-scale 3D reconstruction problem turns into the problem of aligning all the scene fragments. To construct a scene fragment $\mathbf{F}_i \in \mathcal{F}$, we simply use KinectFusion~\cite{Newcombe:2011} for every $N$ frames, \eg 50, which is a volumetric approach to reconstruct a scene with truncated signed distance functions~\cite{Curless:1996}. Afterwards, we find pairwise transformations $\mathbf{T}_{i,i+1}$ for all pairs of the consecutive fragments and align all the fragments in the world coordinate system based on sequential multiplication of the pairwise transformations. 

\vspace{1mm}\noindent\textbf{Loop closure detection:} The set of the registered fragments via the sequential multiplication of the pairwise transformations usually has a large amount of accumulated pose errors as well as misaligned range data. Therefore, it is necessary to identify loop closures to diffuse drift errors across all the fragments. To detect loop closures, we align all pairs of the inconsecutive fragments using the FPFH descriptor~\cite{Rusu:2009} and check the overlap ratio of the aligned fragments. If the overlapping ratio between the fragments $\mathbf{F}_i$ and $\mathbf{F}_j$ exceeds a predefined percentile, \eg 30\%, we determine the fragment pair as a loop closure and define its pairwise transformation as $\mathbf{T}_{i,j}$.

\vspace{1mm}\noindent\textbf{Pose graph optimization:} Given a set of loop closures, we minimize the drift errors through the pose graph optimization. Here, we adopt the line process as in~\cite{Choi:2015} to handle spurious loop closures obtained by low distinctiveness of 3D local descriptors.
For a set of fragments $\mathcal{F} = \{  \mathbf{F}_0, ..., \mathbf{F}_n \}$, we define a set of transformations $\mathcal{T} = \{ \mathbf{T}_0, ..., \mathbf{T}_n \}$ where $\mathbf{T}_i$ is a transformation from a fragment $\mathbf{F}_i$ to world reference coordinates and a pairwise transformation from $\mathbf{F}_j$ to $\mathbf{F}_i$ is expressed as $\mathbf{T}_{i,j} = \mathbf{T}_i^{-1} \circ \mathbf{T}_j$. 
Then, given pairwise transformations $\mathbf{T}_{i,j}$ between the fragments $\mathbf{F}_i$ and $\mathbf{F}_j$, we estimate the transformations $\mathcal{T}$ of the fragments $\mathcal{F}$ and a line process $\mathcal{L}$ by minimizing the following function, 
%
\begin{equation}
\vspace{-2pt}
\begin{split}
	E_L(\mathcal{T},\mathcal{L}) &= \sum_{i} f(\mathbf{T}_i, \mathbf{T}_{i+1}, \hat{\mathbf{T}}_{i,i+1}) \\ 	&+\sum_{i,j} l_{ij} f(\mathbf{T}_i, \mathbf{T}_j, \hat{\mathbf{T}}_{i,j})  
	+ \sum_{i,j} \Psi(l_{ij}),
\end{split}\label{eqs:posegraph}
\vspace{-2pt}
\end{equation}
where $f(\mathbf{T}_a,\mathbf{T}_b,\hat{\mathbf{T}}_{a,b})$ measures the difference between the pre-computed pairwise transformation $\hat{\mathbf{T}}_{a,b}$ and the pairwise transformation computed from $\mathbf{T}_a$ and $\mathbf{T}_b$. $l_{ij} \in \mathcal{L}$ is a parameter of a line process. $\Psi(l_{ij}) = \sqrt{1-l_{ij}^2}$ is a constraint to maximize the number of inlier loop closures. If an estimated parameter $l_{ij}$ is larger than a threshold, the loop closure between the fragments $i$ and $j$ is determined as a correct loop closure.

\subsection{Layout Estimation} \label{sec:layoutest}

To estimate the scene layout, which consists of a set of planes such as a ceiling, a floor, and walls, we find the dominant planes $\mathcal{P}_\text{dominant}$ in the scene and then determine layout planes $\mathcal{P}_\text{layout}$ from $\mathcal{P}_\text{dominant}$. To extract a set of  dominant planes, we compute and cluster plane parameters from supervoxels~\cite{Papon:2013} of each fragment and subsequently merge similar plane parameters in the world coordinate system.

\begin{figure}[t]
	\centering \hspace{-1.0mm}
	\subfigure[Clustered planes]{
		\includegraphics[width=0.45\columnwidth]{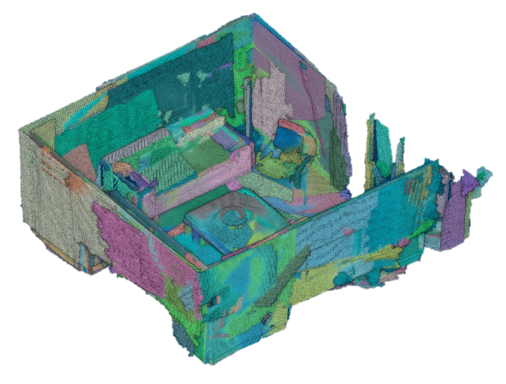}\label{fig:layoutEstimation_sub1}
	}\subfigure[Dominant planes]{
		\includegraphics[width=0.45\columnwidth]{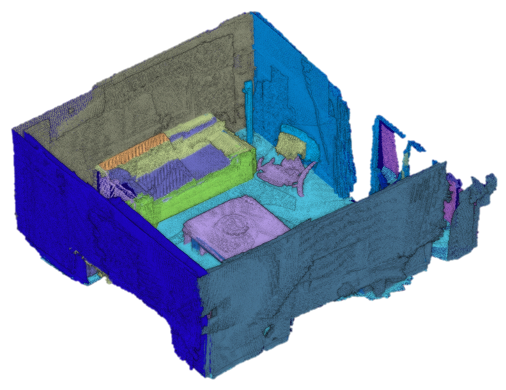}\label{fig:layoutEstimation_sub2}
	}\caption{The result of the hierarchical agglomerative clustering and energy-based multi-model fitting. This procedure approximates the scene with the a small number of planes. Therefore, it is easy to find the scene layout from these planes.}\label{fig:layoutEstimation}\vspace{-10.0pt}
\end{figure}

\vspace{1.0mm}\noindent\textbf{Dominant plane extraction:}
Initially, we divide a fragment $\mathbf{F}_i$ into a set of supervoxels~\cite{Papon:2013}, $\mathcal{S} = \{ \mathcal{S}_1 \cup \mathcal{S}_2 \cup ... \cup \mathcal{S}_K \}$, and generate plane hypotheses using the supervoxels. To generate a plane hypothesis $\pi_l$ in the fragment $\mathbf{F}_i$, we compute a plane parameter from the center points of three adjacent supervoxels because it improves computational efficiency in comparison with the way that a plane parameter is computed using all the points in a sampled supervoxel. Here, the number of initial plane hypotheses proportionally increases as the scale of a scene increases, and there might be a lot of similar planes owing to largely planar regions such as walls. Therefore, we cluster initial plane hypotheses through two plane clustering steps. First, we merge the plane hypotheses using the hierarchical agglomerative clustering~\cite{Magri:2014}. The distance between a supervoxel $\mathcal{S}_k$ and a plane hypothesis $\pi_l$ is computed as
%
\begin{equation}
\vspace{-4pt}
\label{eqs:plane2supervoxel_dist}
C(\pi_l,\mathcal{S}_k) = \frac{1}{|\mathcal{S}_k|}\sum_{\mathbf{p} \in \mathcal{S}_k} d(\pi_l, \mathbf{p}).
\vspace{-0pt}
\end{equation}
The distance function $d(\cdot, \cdot)$ is defined as
%
\begin{equation}
\vspace{-2pt}
\label{eqs:plane2point_dist}
d(\pi_l, \mathbf{p}) = \dfrac{|\pi_l^\top \mathbf{\bar{p}}|}{\sqrt{a_l^2 + b_l^2 + c_l^2}}, 
\vspace{-2pt}
\end{equation}
where a plane parameter $\pi_l$ is denoted as $\pi_l = \left[ a_l, b_l, c_l, 1 \right]^\top$ and $\mathbf{\bar{p}}$ is a homogeneous representation of a 3D point $\mathbf{p}$. Some supervoxels with similar plane hypotheses are grouped together via the clustering method and used to recalculate plane parameters. However, there still exist some plane hypotheses that are on an identical wall but not grouped together because of local distortion in the vicinity of the fragment's border. Thus, as the second step, we assign the recomputed plane hypotheses to each 3D point by minimizing an energy function via graph cuts~\cite{Boykov:2001}. Given a set of 3D points, \ie, $\mathbf{F}_i$, and a set of plane parameters, denoted by $\mathcal{P}_i$, the problem is defined as finding a mapping function $h$ from a point $\mathbf{p} \in \mathbf{F}_i$ onto a plane parameter $\pi_l \in \mathcal{P}_i$ (\ie, $h$: $\mathbf{F}_i \mapsto \mathcal{P}_i$). An energy function $E_P$ is defined as
%
\begin{equation}
\label{eqs:graphcut_energy_func}
E_P(h) = \sum_{\mathbf{p}\in\mathbf{F}_i} D_\mathbf{p}(h_\mathbf{p}) + \sum_{\mathbf{p}\in\mathbf{F}_i,\mathbf{q}\in\mathcal{N}_\mathbf{p}} V_{\mathbf{p},\mathbf{q}}(h_\mathbf{p}, h_\mathbf{q}).
\end{equation}
The data term $D_\mathbf{p}$ is defined in the same manner as Eq.~(\ref{eqs:plane2point_dist}) to measure the distance between a point $\mathbf{p}$ and a plane parameter $\pi_l$. We employ the Potts model~\cite{Boykov:2001} as the smoothness term $V_{\mathbf{p},\mathbf{q}}$ to preserve continuity of a plane parameter between neighboring points. The Potts model is defined as $V_{\mathbf{p},\mathbf{q}}(h_\mathbf{p},h_\mathbf{q}) = \alpha_{\mathbf{p},\mathbf{q}} T(h_\mathbf{p} \neq h_\mathbf{q})$ where $\alpha_{\mathbf{p},\mathbf{q}}$ is a penalty weight and $T$ is 1 if the argument is true and otherwise 0. $\mathcal{N}_\mathbf{p}$ represents neighboring points of $\mathbf{p}$. The neighboring points are determined as points within a predefined distance among points obtained by the $k$-nearest neighbor ($k$-NN) search algorithm~\cite{Rusu:2011}. In addition, we employ a null-plane hypothesis $\pi_\emptyset$ to avoid assigning plane hypotheses to a point that has a large displacement from the plane. Therefore, the data term is redefined as
%
\begin{equation}
D_\mathbf{p}(h_\mathbf{p}) = 
\left\{
\begin{array}{lc}
d(h_\mathbf{p}, \mathbf{p}), & \mathrm{if}\ h_\mathbf{p} \neq \pi_\emptyset \\
\gamma, & \text{otherwise} \\
\end{array},
\right. 
\label{eqs:dataterm_for_graphcut}
\end{equation}
where $\gamma$ is a constant. Here, the role of the null hypothesis is to ignore noisy measurements or points on non-planar surfaces. As a result, we obtain a smaller number of merged plane hypotheses. The plane hypotheses in each fragment $\mathbf{F}_i$ are transformed from the fragment coordinate system to the world reference coordinate system. Example of clustered planes are shown in Fig.~\ref{fig:layoutEstimation_sub1}.
%

With the clustered and transformed plane hypotheses, we find a set of dominant planes $\mathcal{P}_\text{dominant}$, which enables to represent the scene with a small number of plane hypotheses. To find $\mathcal{P}_\text{dominant}$, we employ the hierarchical agglomerative clustering again. 
Here, instead of comparing supervoxels, we compute the distance function of Eq.~(\ref{eqs:plane2supervoxel_dist}) using groups of 3D points with the same plane hypothesis via Eq.~(\ref{eqs:graphcut_energy_func}).
Consequently, it is possible to acquire the set of planes that best describe the scene as shown in Fig.~\ref{fig:layoutEstimation_sub2}, where different colors indicate that different plane hypotheses are assigned. 

\vspace{1.0mm}\noindent\textbf{Layout plane estimation:} Given dominant planes $\mathcal{P}_\text{dominant}$ and clustered point clouds, we estimate the scene layout which can be understood as an envelope of an arbitrary indoor space that includes the ceiling, floor, and walls. Here, we assume a weak Manhattan world in which all the walls are orthogonal to the ceiling and the ground floor, but the walls are not necessarily orthogonal to each other. However, in practice, captured planes are hard to be perfectly planar due to the measurement noise, and therefore, we make planes orthogonal to each other if they are quite close to be orthogonal. 

We find the scene layout planes $\mathcal{P}_\text{layout}$ through two steps. In the first step, we find the ceiling or ground floor, called a base plane, assuming that one of them is the largest planar region among all the plane hypotheses. The base plane is determined by computing the areas of dominant planes instead of simply counting the number of clustered 3D points because the density of 3D points significantly differs depending on the amount of acquired range data. To estimate the area of a dominant plane, we generate a 2D occupancy grid map on the dominant plane and project labeled 3D points on to the dominant plane. Then, we count the number of occupied grids.

\begin{figure}[t]
	\centering \hspace{-1.0mm}
    \includegraphics[width=0.95\columnwidth]{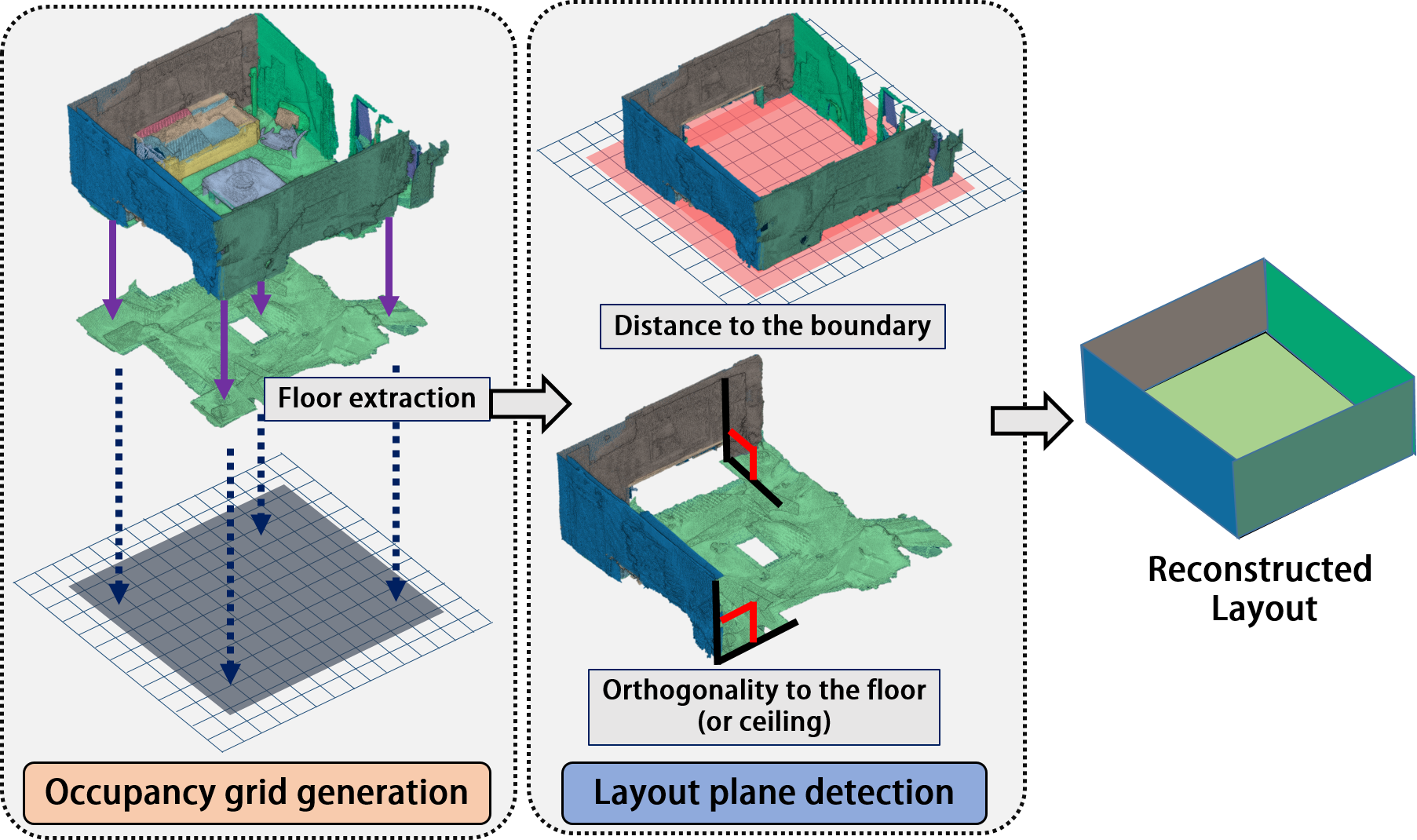}
    \caption{Layout estimation procedure. We extract a floor (or a ceiling) and generate a 2D occupancy grid. Afterwards, we find a set of layout planes by checking two criteria, boundary distance and orthogonality.}\label{fig:grid}\vspace{-15.0pt}
\end{figure}

In the second step, we find a set of planes that are orthogonal to the base plane determined in the first step as follows. We generate a 2D occupancy grid map on the base plane as shown in Fig.~\ref{fig:grid}. Then, we project all the 3D points onto the base plane and fill each cell of the grid map if the density of points is larger than a predefined value. Here, an empty cell indicates that it is either outside the room or inside the room but not measured. From the occupancy grid, we determine the boundary of occupied grids, denoted by $\partial O$, via the morphological boundary detection~\cite{Iwanowski:2007} that can handle an arbitrary shape. Finally, we select the set of planes by following criteria:
%
\begin{equation}
L(\pi_i) = 
\left\{
\begin{array}{lc}
\text{1}, & \mathrm{if}\ (\vec{n}_i \cdot \vec{n}_\text{base}) < \tau_1\ \text{and}\ g(\partial O, \pi_i) < \tau_2 \\
\text{0}, & \text{otherwise} \\
\end{array},
\right.
\label{eqs:layoutplane_criteria}
\end{equation}
where $\vec{n}_i$ and $\vec{n}_\text{base}$ are normal vectors of a selected dominant plane and base plane, respectively. Therefore, the first criterion checks the perpendicularity between two planes. The second criterion checks the distance between the plane $\pi_i$ and the boundary $\partial O$ because the layout planes, especially walls, surround the space. $\tau_1$ and $\tau_2$ are two user-defined parameters. The distance function $g$ is defined as
%
\begin{equation}
\label{eqs:boundary2plane_dist}
g(\partial O, \pi_i)= \frac{1}{|\mathcal{S}_{\pi_i}|} \sum_{\mathbf{p} \in \mathcal{S}_{\pi_i}}|\partial O - \mathbf{p}_\text{proj}|, 
\end{equation}
where $\mathbf{p}_\text{proj}$ is the projected point of $\mathbf{p}$ on to the base plane and $\mathcal{S}_{\pi_i}$ denotes a set of 3D points that belong to the plane $\pi_i$. $|\mathcal{S}_{\pi_i}|$ is the cardinality of $\mathcal{S}_{\pi_i}$. If two criteria are satisfied, we regard the corresponding dominant plane as a layout plane. Figure~\ref{fig:grid} shows the result of layout plane estimation. 
%

\subsection{Global Registration with Scene Layout} \label{sec:globalreg}

As the last step, we reconstruct the entire scene by registering all the fragments $\mathcal{F}$ with the aid of the estimated scene layout in the world coordinate system. We pose a global optimization problem for the layout-constrained global registration. To resolve the problem, we introduce a joint approach of the layout estimation and the global registration because they depend on each other. A detailed description is given in the following subsections.

\vspace{1.0mm}\noindent\textbf{Terminologies:}
Let $\mathcal{I}$ denote a set of fragment pairs. If a pair of fragments, $(\mathbf{F}_i,\mathbf{F}_j) \in \mathcal{I}$, has an overlapping region, we define a set of correspondences, $\mathcal{C}_{i,j}$, between their points in the overlapping region. Similarly, we define a set of correspondences, $\mathcal{C}_i$, between each fragment $\mathbf{F}_i$ and the layout where the corresponding point of the layout is a virtual point on the layout plane. For example, we project a point $\mathbf{p}$ of a fragment $\mathbf{F}_i$ onto the nearest layout plane and establish a correspondence if the distance between the point and the virtual (projected) point is small. For a point $\mathbf{p} \in \mathbb{R}^3$ and a transformation $\mathbf{T}$, a transformed point is represented by $\mathbf{T}(\mathbf{p}) = \mathbf{R} \mathbf{p} + \mathbf{t}$ where $\mathbf{R} \in SO(3)$ is a rotation matrix and $\mathbf{t} \in \mathbb{R}^3$ is a translation vector.

\vspace{1.0mm}\noindent\textbf{Objective function:} For the global registration of all fragments, we define the following energy function,
%
\begin{equation}
\begin{split}
	E_R(\mathcal{T}) &= \sum_{i} E_\text{layout}(\mathbf{T}_i) + \lambda_1 \sum_{i, j}E_\text{frag}(\mathbf{T}_i, \mathbf{T}_j)  \\
	&+ \lambda_2 \sum_{i, j} E_\text{pair}(\mathbf{T}_i, \mathbf{T}_j),
\end{split}\label{eqs:globalreg}
\end{equation}
where $\lambda_1$ and $\lambda_2$ are weighting parameters and are determined depending on the numbers of points and fragment pairs. 
The first term is to minimize the distance between correspondence points of the layout and each fragment.
Among many metrics~\cite{Besl:1992,Chen:1992,Liu:2014}, we use the point-to-plane metric~\cite{Chen:1992}. By the metric, the first term is defined as
%
\begin{equation}
\label{eqs:layoutcost}
	E_\text{layout}(\mathbf{T}_i) = \sum_{(\mathbf{p},\mathbf{q}) \in \mathcal{C}_{i}} \left \| (\mathbf{T}_i(\mathbf{p}) - \mathbf{q})^\top \mathbf{R}_i \mathbf{n}_{\mathbf{p}} \right \| ^2,
\end{equation}
where $\mathbf{n}_{\mathbf{p}}$ is a normal vector of $\mathbf{p}$ and $\mathbf{q}$ is a virtual (projected) point on the layout. Since the layout is estimated under the weak Manhattan world assumption in Sec. \ref{sec:layoutest}, the aligned fragments have axis-aligned geometry, \eg, orthogonality between a wall and the ceiling. It is worthy of note that, since we only constrain the positions of points along the envelope of the scene, objects inside the space are not necessarily planar.
The second term is to minimize the distance between correspondence points of each pair of fragments.
In the same way as the point-to-plane metric~\cite{Chen:1992}, the second term is defined as
%
\begin{equation}
\label{eqs:pointcost}
	E_{\text{frag}}(\mathbf{T}_i, \mathbf{T}_j) = \sum_{(\mathbf{p},\mathbf{q}) \in \mathcal{C}_{i,j}} \left \| (\mathbf{T}_i(\mathbf{p}) - \mathbf{T}_j(\mathbf{q}))^\top \mathbf{R}_i \mathbf{n}_{\mathbf{p}} \right \| ^2.
\end{equation}
For the last term, we incorporate the pairwise transformation constraint as
%
\begin{equation}
\label{eqs:paircost}
	E_\text{pair}(\mathbf{T}_i, \mathbf{T}_j) = \delta \left( \mathbf{T}_i \circ \tilde{\mathbf{T}}_{i,j} - \mathbf{T}_{j} \right) , 
\end{equation}
where $\tilde{\mathbf{T}}_{i,j}$ is a pairwise transformation estimated by the iterative closest point (ICP) algorithm~\cite{Chen:1992} and $\delta$ is the sum of the norms of elements. The pairwise transformation constrains the feasible solution space to avoid a degenerate situation, \eg, a fragment moves too much or the scene structures are collapsed. To optimize Eq.~(\ref{eqs:globalreg}), we use the widely known Gauss-Newton method. 

\vspace{1.0mm}\noindent\textbf{Joint optimization:} Since the layout estimation and the global registration problems are closely related to each other, we alternately estimate the scene layout and the optimal transformations instead of solving the complex joint estimation problem. Algorithm~\ref{alg:globalopt} shows the entire procedure. Initially, we regard that the initial transformations $\mathcal{T}_0$ and fragments $\mathcal{F}$ are given. Here, we set the coordinates of the first fragment to the world reference coordinates so that $\mathbf{T}_0$ is fixed to an identity matrix. Afterwards, we repeatedly estimate the scene layout and minimize the objective function in Eq.~(\ref{eqs:globalreg}). We experimentally confirmed that the inner and outer loops in Algorithm~\ref{alg:globalopt} generally converge commonly within 10 and 20 iterations, respectively.
%
\begin{algorithm} [tb] 
\caption{Joint layout estimation and global registration}
\label{alg:globalopt}
\begin{algorithmic}[1]
\REQUIRE $\mathcal{F}$, $\mathcal{T}_{0}$
\ENSURE $\mathcal{T}$
	\STATE establish $\mathcal{C}_{i,j}$ $\forall (\mathbf{F}_i,\mathbf{F}_j) \in \mathcal{I}$ 
	\STATE $\mathcal{T} \leftarrow \mathcal{T}_{0}$
	\REPEAT
		\STATE estimate $\mathcal{P}_\text{layout}$ using the method of Sec.~\ref{sec:layoutest}
		\STATE establish $\mathcal{C}_{i}$ $\forall \mathbf{F}_i \in \mathcal{F}$
		\REPEAT
			\STATE compute $\Delta \mathcal{T}$ using Eq.~(\ref{eqs:globalreg})
            \STATE $\mathcal{T} \leftarrow \mathcal{T} + \Delta \mathcal{T}$
		\UNTIL $N$ times
		\STATE transform $\mathcal{F}$ using $\mathcal{T}$
	\UNTIL $M$ times
\end{algorithmic} 
\end{algorithm} 

\begin{figure}[tb]
	\begin{center}
		\begin{tabular}{@{}c@{}}
   		    \includegraphics[width=0.95\columnwidth,height=77pt]{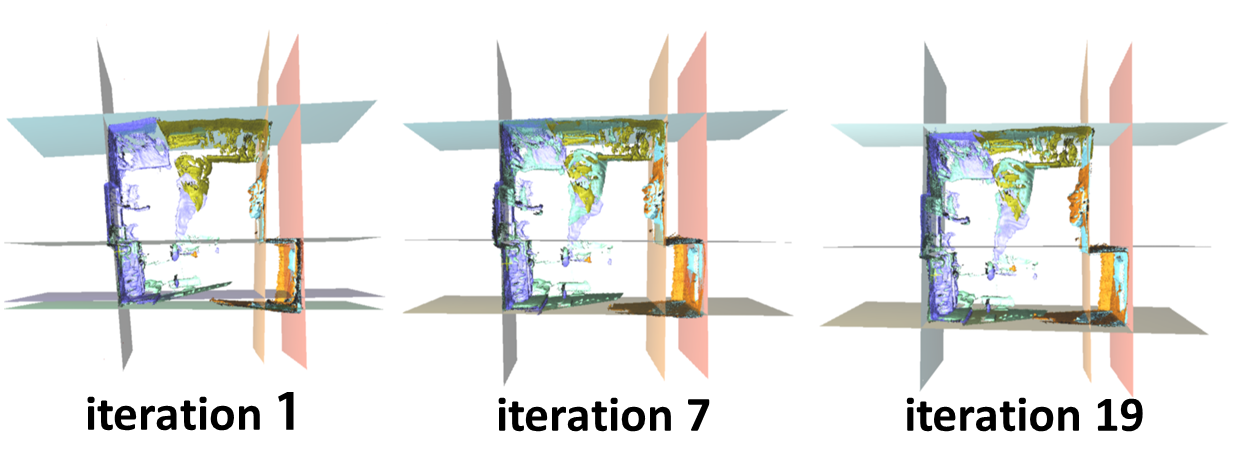}
		\end{tabular}
	\end{center} \vspace{-10pt}
	\caption{Joint optimization procedure. As we iterate layout estimation and global registration procedures, the fragmented structures merge into a wall region.} \label{fig:iterativeopt} \vspace{-12pt}
\end{figure}

The optimized process is shown in Fig.~\ref{fig:iterativeopt}. As the number of the iterations increases, the curved walls are straightened more. The reconstructed room has a cuboid shape at the end of the iterations. Consequently, the joint approach improves the global registration and the layout estimation. 
%

\section{Experimental Results}

We experimentally verified the proposed method in quantitative and qualitative ways by using publicly available datasets: the augmented ICL-NUIM dataset~\cite{Choi:2015} and the SUN3D dataset~\cite{Xiao:2013}. The former is a synthetic dataset generated in consideration of a noise model of a consumer depth camera. Since this dataset provides the ground truth trajectories and 3D structures, we performed the quantitative evaluation using this dataset. In contrary, the SUN3D dataset was captured in the real-world environment using a hand-held camera and did not provide the ground truth information. Thus, we use this dataset to confirm the feasibility of the proposed method in practical situations. For evaluation, we compare the proposed method with the state-of-the-art methods~\cite{Whelan:2012,Whelan:2015,Xiao:2013,Choi:2015}. Here, Kintinuous~\cite{Whelan:2012} and ElasticFusion~\cite{Whelan:2015} are online methods, and SUN3D structure from motion (SFM)~\cite{Xiao:2013}, the Choi \etal method~\cite{Choi:2015}, and the proposed method are offline methods. \emph{Please note that detailed parameters used for our experiments and more results can be found in the supplementary material.}

\begin{table}[t]
\caption{Reconstruction performance evaluation in terms of average and median errors by using four synthetic datasets. The unit of error is centimeter. The best performance in each row is represented in bold.} \vspace{3pt}
\label{tab:reconstruction_error}\centering
\setlength{\tabcolsep}{3.5pt}
\footnotesize
\begin{tabular}{c|c|c|c|c|c|c}\hline\hline
\multicolumn{2}{c|}{}  		& Kint.~\cite{Whelan:2012} & Elas.~\cite{Whelan:2015} & SUN3D~\cite{Xiao:2013} & Choi~\cite{Choi:2015} & Ours \\\hline
\multirow{2}{*}{\emph{Liv.1}}	& Avg.		& 	13.19	&	9.31	&	12.69	&	5.41	& 	\textbf{2.72} \\\cline{2-7}
									& MED		&	7.47 	&	4.96	&	5.85	& 	4.39 	& 	\textbf{1.56} 	\\\hline
\multirow{2}{*}{\emph{Liv.2}}	& Avg.		&	11.60 	&	12.11	&	10.53	& 	7.12 	& 	\textbf{5.43} 	\\\cline{2-7}
									& MED		&	7.45 	&	6.41	&	5.79	& 	3.65 	& 	\textbf{3.25} 	\\\hline
\multirow{2}{*}{\emph{Off.1}}	& Avg.		&	9.01 	&	4.89	&	34.41	& 	\textbf{3.51}	&	4.02	\\\cline{2-7}
									& MED		&	5.75 	&	2.67	&	28.04	& 	\textbf{2.64} 	&	2.72	\\\hline
\multirow{2}{*}{\emph{Off.2}}	& Avg.		&	9.48 	&	5.36	&	33.09	& 	3.52 	& 	\textbf{3.14}	\\\cline{2-7}
									& MED		&	4.33 	&	2.30	&	29.61	& 	1.92 	& 	\textbf{1.79}	\\\hline\hline
\end{tabular} \vspace{0pt}
\end{table} 

\vspace{1.0mm}\noindent\textbf{Reconstruction quality:} To measure the quality of estimated structures, we compute the average and the median of errors. The errors are defined as the distance between an estimated 3D point and its closest ground truth point. Table~\ref{tab:reconstruction_error} shows the reconstruction errors of the proposed method and those of the-state-of-the-art methods. Overall, the proposed method shows superior results compared to the state-of-the-art methods, except the \emph{Office1} dataset. Occasionally, the reconstruction error of the proposed method is slightly higher than that of the Choi \etal method because of overfitting noisily reconstructed fragments to the scene layout. However, in most cases, the proposed method shows more accurate results than the Choi \etal method because planar structures are preserved better than the results of the Choi \etal method. 

\begin{figure}[tb]
	\begin{center}
		\begin{tabular}{@{}c@{}c@{}c@{}}
   		    \includegraphics[width=0.05\columnwidth]{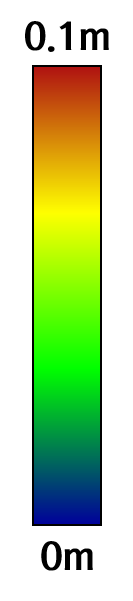} \ &
			\includegraphics[width=0.47\columnwidth]{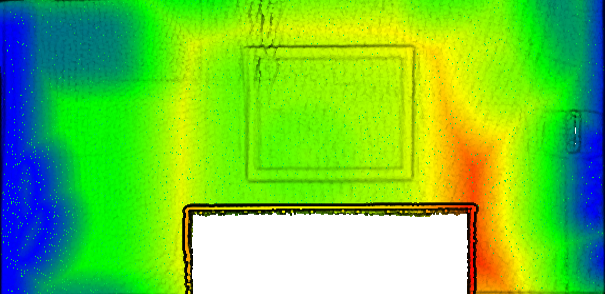} \ &
			\includegraphics[width=0.47\columnwidth]{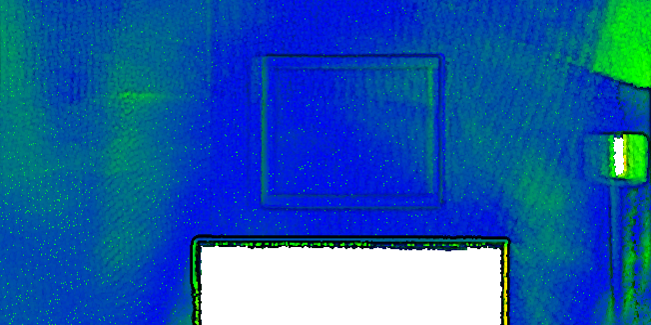} \\
			& (a) & (b) \\ \vspace{-9pt}
		\end{tabular}
	\end{center} \vspace{-8pt}
	\caption{Visualization of reconstruction errors of the Choi \etal method~\cite{Choi:2015} (a) and the proposed method (b) in a wall region of the \emph{Livingroom1} dataset. The proposed method shows consistently lower errors than (a), with the aid of layout information.}\label{fig:syntheticlayout} \vspace{-10pt}
\end{figure}

\begin{figure*}[t]
	\hspace{-3.0mm}
	\vspace{-5.0mm}
    \centering 
    \subfigure[]{
		\includegraphics[width=0.4\columnwidth]{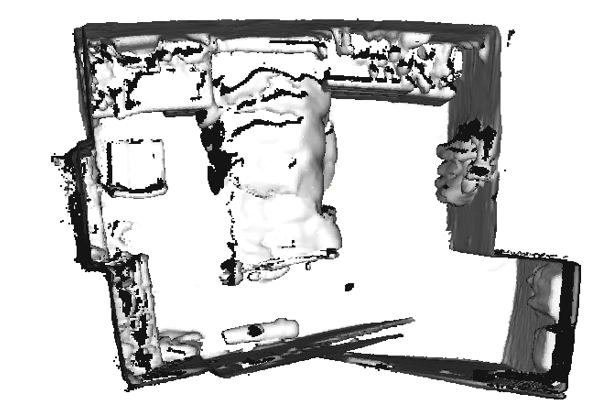}
        \label{fig:iterativeopt_a}
	}\subfigure[]{
		\includegraphics[width=0.33\columnwidth]{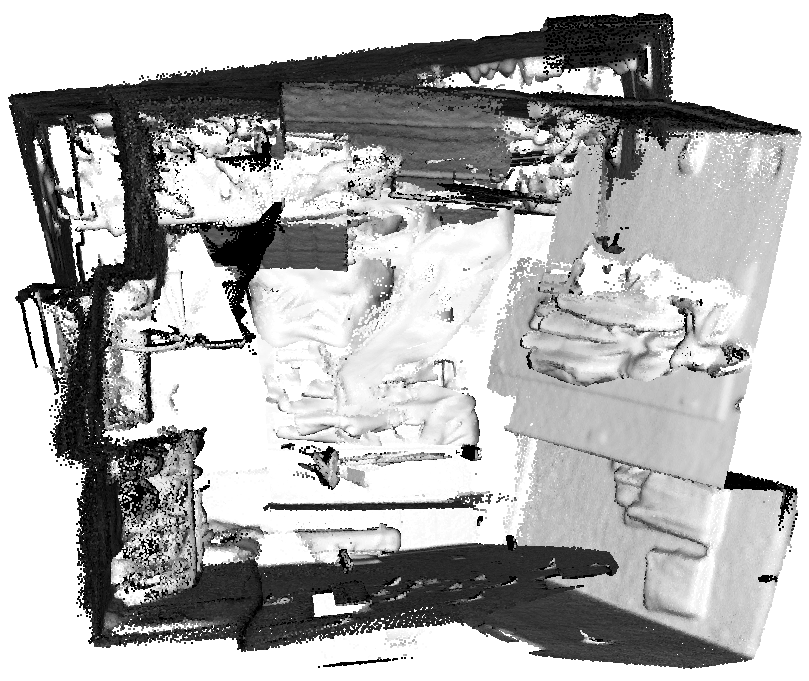}
        \label{fig:iterativeopt_b}
	}\subfigure[]{
		\includegraphics[width=0.4\columnwidth]{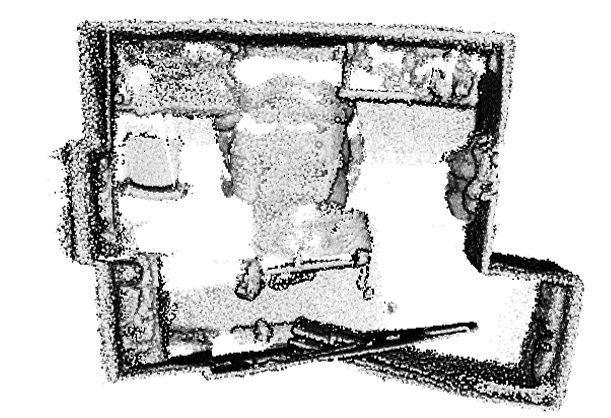}
        \label{fig:iterativeopt_c}        
	}\subfigure[]{
		\includegraphics[width=0.4\columnwidth]{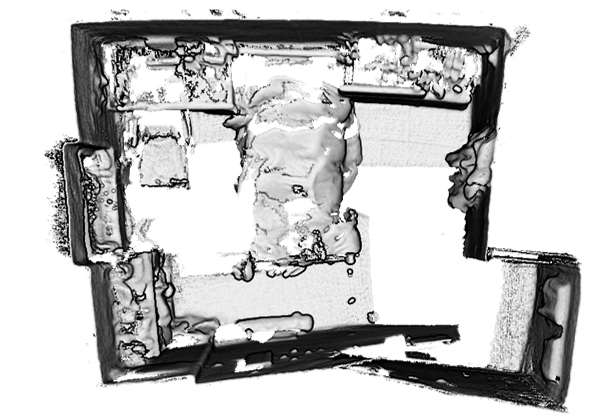}
        \label{fig:iterativeopt_d}
	}\subfigure[]{
		\includegraphics[width=0.4\columnwidth]{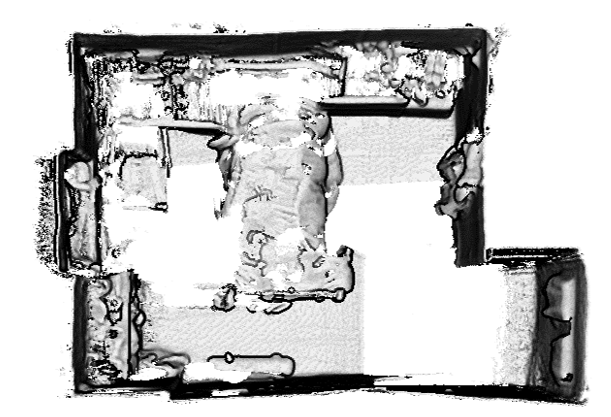}
        \label{fig:iterativeopt_e}
	}\\ \vspace{+3pt}
    \subfigure[]{
		\includegraphics[width=0.4\columnwidth]{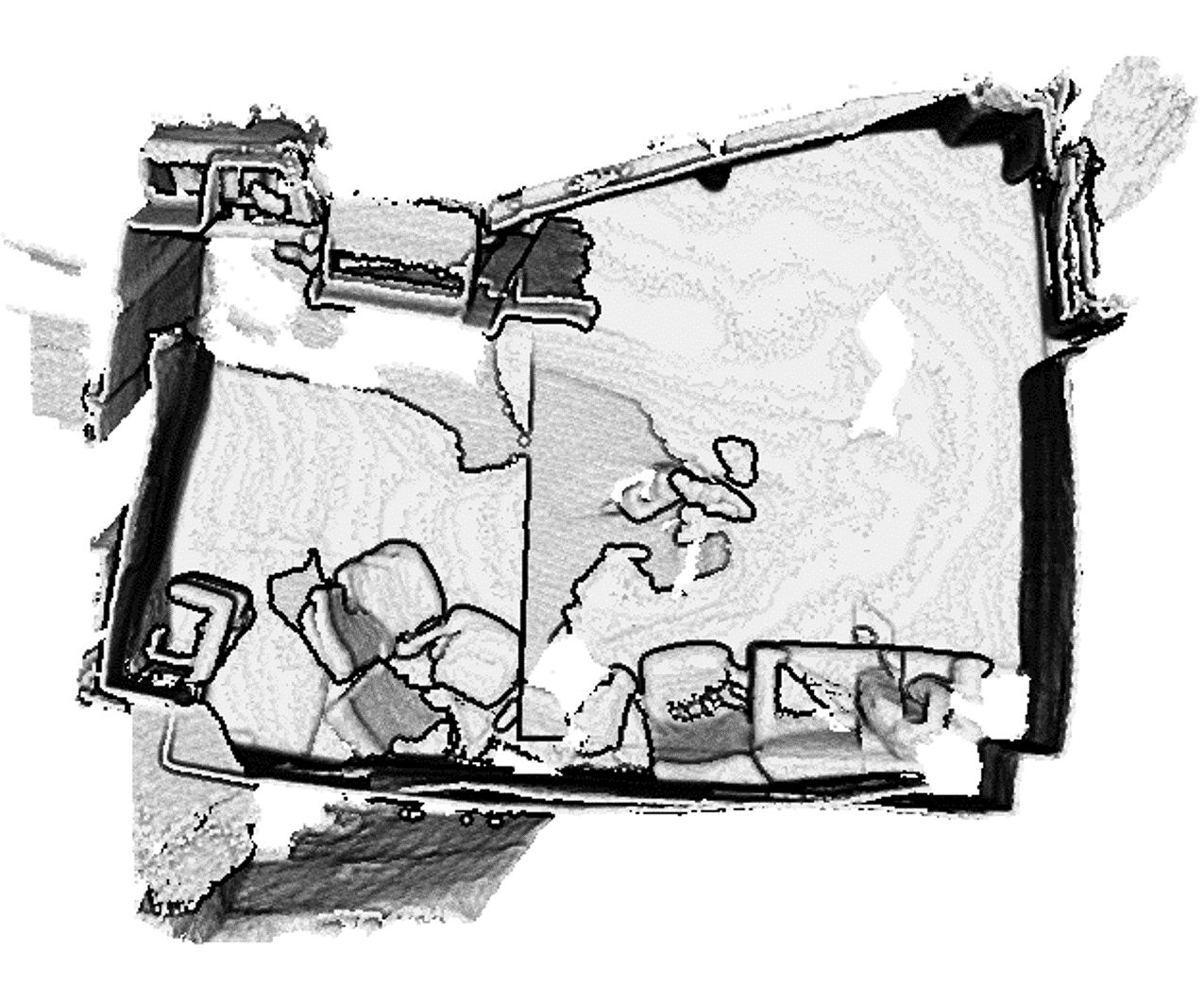}
        \label{fig:iterativeopt_f}
	}\subfigure[]{
		\includegraphics[width=0.34\columnwidth]{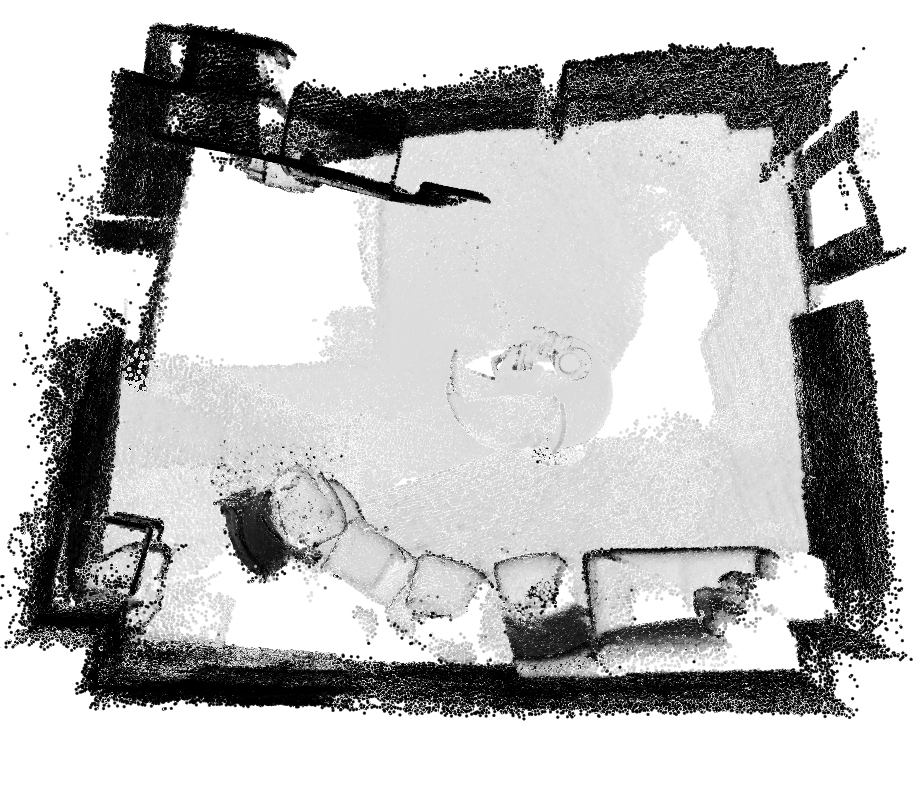}
        \label{fig:iterativeopt_g}		
	}\subfigure[]{
		\includegraphics[width=0.4\columnwidth]{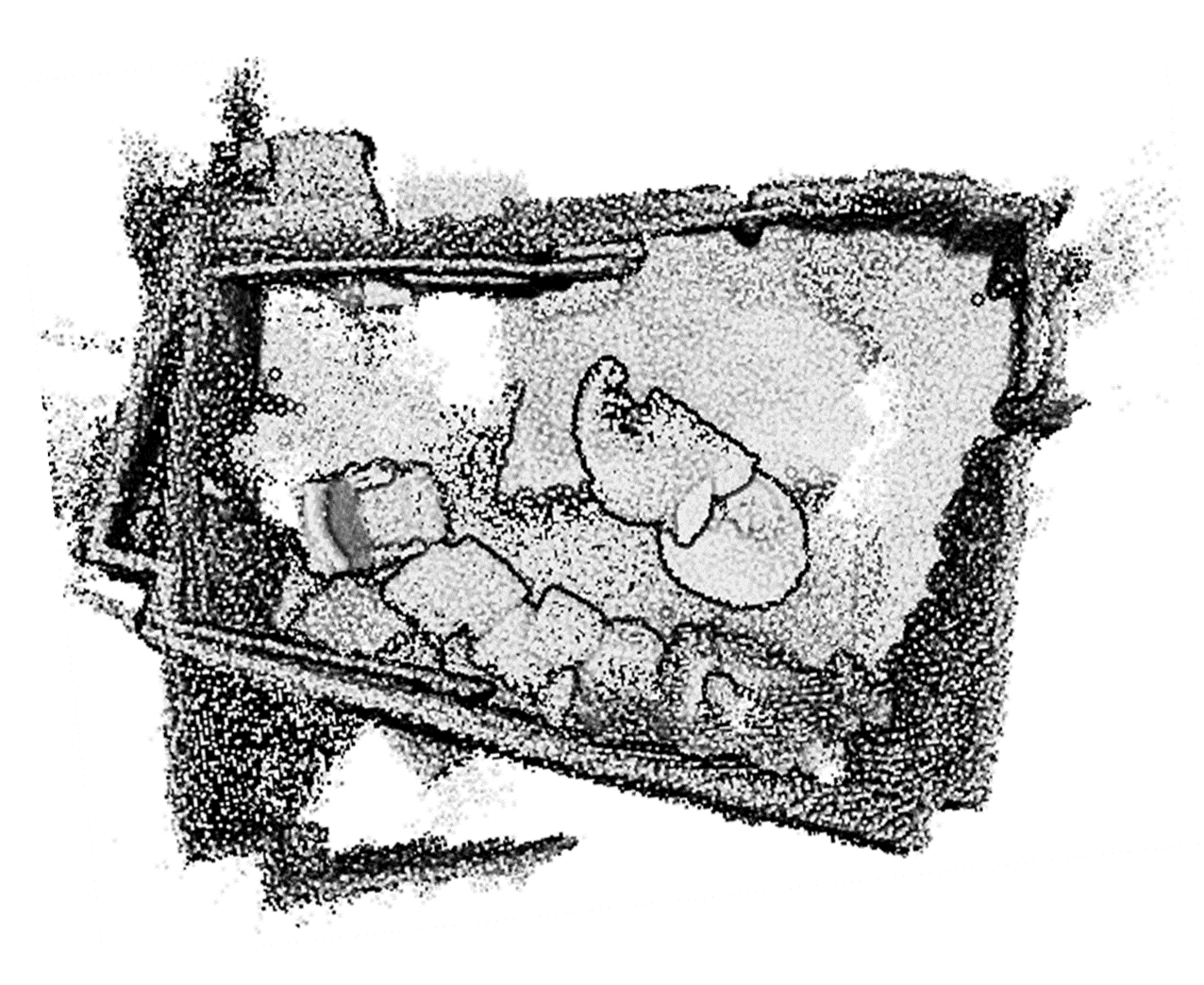}
        \label{fig:iterativeopt_h}		
	}\subfigure[]{
		\includegraphics[width=0.4\columnwidth]{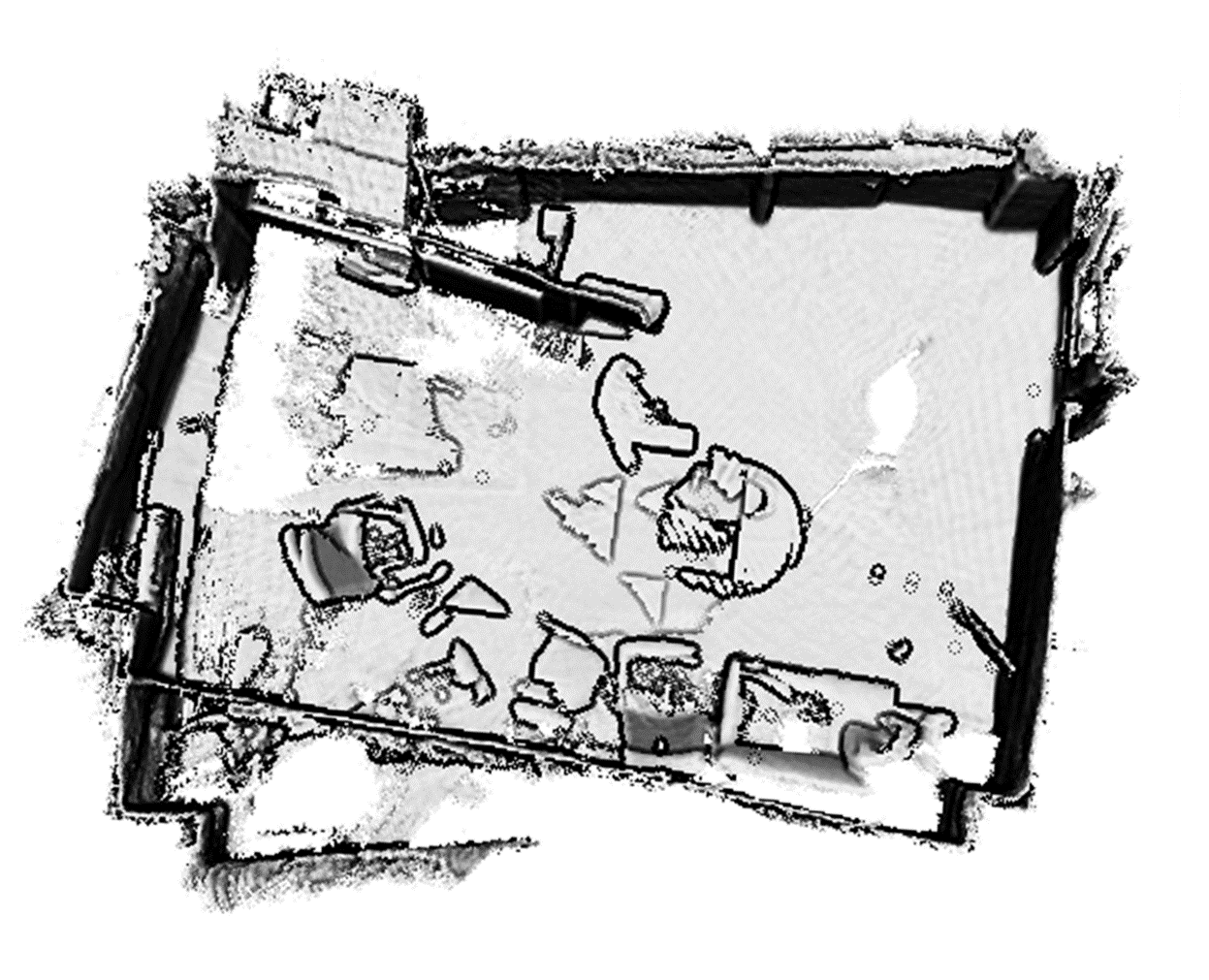}
        \label{fig:iterativeopt_i}		
	}\subfigure[]{
		\includegraphics[width=0.4\columnwidth]{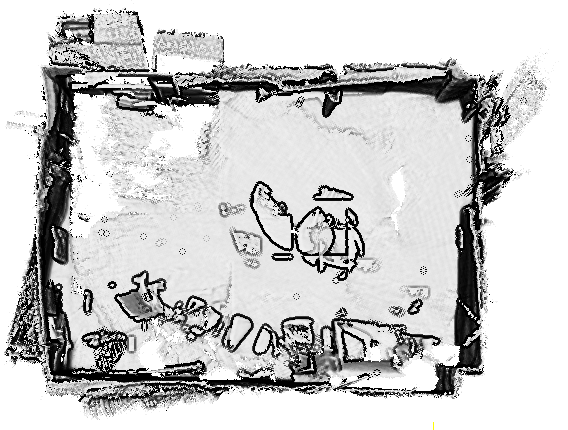}
        \label{fig:iterativeopt_j}		
	}\\ \vspace{+3.0pt}
    \caption{Comparison of reconstructed results for real-world datasets~\cite{Xiao:2013}. From the left to the right; Kintinuous~\cite{Whelan:2012}, ElasticFusion~\cite{Whelan:2015}, SUN3D SFM~\cite{Xiao:2013}, Choi \etal~\cite{Choi:2015}, and the proposed method, respectively. From the top to the bottom; \emph{mit\_dorm\_next\_sj} and \emph{mit\_lab\_hj} datasets, respectively. Reconstructed results are compared using the top-view to clearly show registration errors in wall regions. }\label{fig:realdata}\vspace{-4.0pt}
\end{figure*}

It is worthy of note that the global registration without the layout information frequently shows bended walls and floors owing to noisy measurements as shown in Fig.~\ref{fig:realdata}. In contrast, the proposed method preserves largely planar structures with the aid of scene layout information. For example, Fig.~\ref{fig:syntheticlayout}\textcolor{red}{(a)} shows non-uniform reconstruction errors in the wall region reconstructed by the Choi~\etal method. The proposed method shows consistently small errors over the wall in Fig.~\ref{fig:syntheticlayout}\textcolor{red}{(b)}. This advantage primarily comes from the layout information. Thus, we claim that the layout information is an important cue for accurate indoor scene reconstruction.

\vspace{1.0mm}\noindent\textbf{Trajectory accuracy:} The trajectory error is measured in terms of the root mean squared error (RMSE) and median error between the ground-truth trajectory and an estimated camera trajectory. Since an accurate camera trajectory implies the accurate registration of scene fragments, we use this metric to evaluate various indoor reconstruction methods. Table~\ref{tab:trajectory_error} shows the trajectory errors of the proposed method and the state of the art methods. The proposed method outperforms other methods except the \emph{Office1} dataset similarly as in Table~\ref{tab:reconstruction_error}. This quantitative comparison also confirms that the proposed method is promising, especially when previous approaches cannot preserve the global scene structures well.

\begin{table}[t]
\caption{Comparison of trajectory errors in terms of root mean squared errors (RMSE) and median errors for synthetic datasets. The errors are measured in centimeter. The best performance in each row is represented in bold.} \vspace{3pt}
\setlength{\tabcolsep}{3.0pt}
\label{tab:trajectory_error}\centering
\footnotesize
\begin{tabular}{c|c|c|c|c|c|c}\hline\hline
\multicolumn{2}{c|}{} 	& Kint.~\cite{Whelan:2012} & Elas.~\cite{Whelan:2015} & SUN3D~\cite{Xiao:2013} & Choi~\cite{Choi:2015} & Ours \\\hline
\multirow{2}{*}{\emph{Liv.1}}	& RMSE	& 57.36		& 59.02		&	32.22	&	9.87	& 	\textbf{9.49} \\\cline{2-7}
									& MED	& 45.16		& 43.83		&	27.28 	&	\textbf{7.88}	& 	8.18 \\\hline
\multirow{2}{*}{\emph{Liv.2}}	& RMSE	& 29.32		& 37.09		&	29.13 	&	13.63	& 	\textbf{12.18}	\\\cline{2-7}
									& MED	& 27.16 	& 24.67		&	24.15	& 	11.81 	&	\textbf{10.50} \\\hline
\multirow{2}{*}{\emph{Off.1}}	& RMSE	& 18.29 	& 13.10		&	50.84	& 	\textbf{6.22}  	& 	9.95	\\\cline{2-7}
									& MED	& 12.11 	& 9.69		&	42.68	& 	\textbf{5.39} 	&	9.31	\\\hline
\multirow{2}{*}{\emph{Off.2}}	& RMSE	& 27.18 	& 13.26		&	29.75	& 	8.89 	& 	\textbf{6.93}	\\\cline{2-7}
									& MED	& 25.25 	& 11.89		&	28.40	& 	9.02 	&  	\textbf{5.86} \\\hline\hline
\end{tabular} \vspace{-10pt}
\end{table} 

%
\begin{figure}[t]
	\begin{center} 
		\subfigure[\cite{Choi:2015} with loop closure]{
   		    \includegraphics[width=0.46\columnwidth]{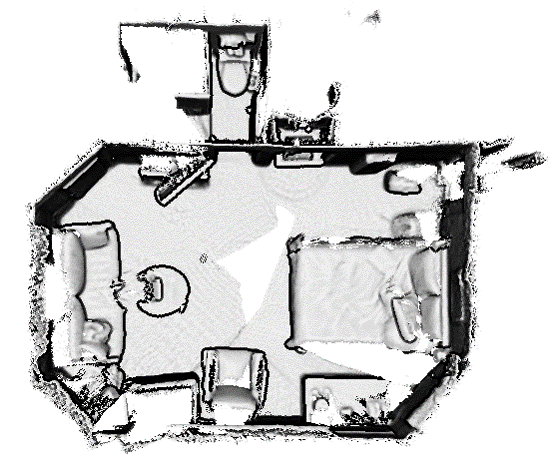}
   		    \label{fig:weakMWexpa}
   		    }
   		\subfigure[Ours with loop closure]{
			\includegraphics[width=0.46\columnwidth]{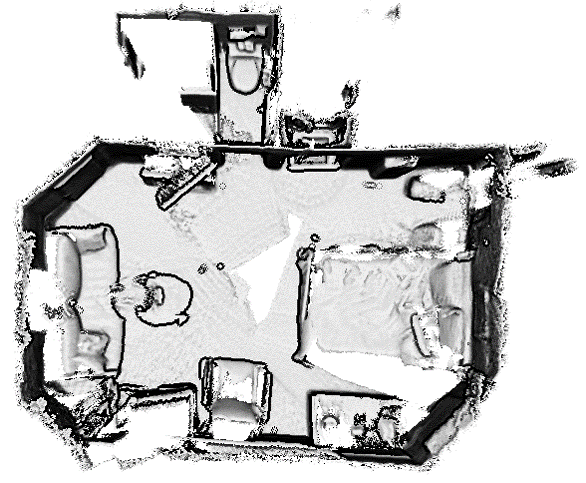}
			\label{fig:weakMWexpb}
			}\\
   		\subfigure[\cite{Choi:2015} without loop closure]{
			\includegraphics[width=0.46\columnwidth]{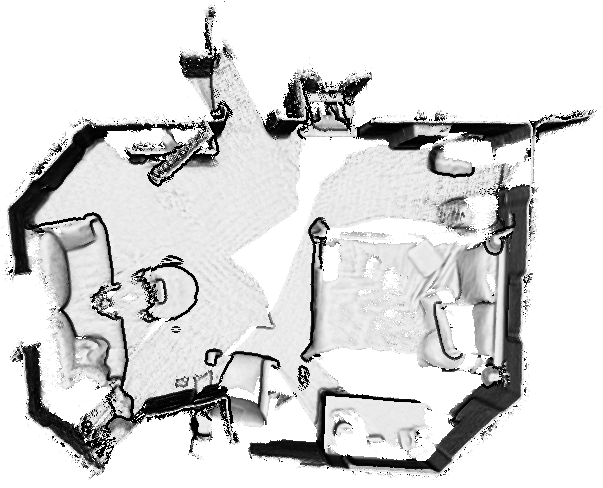}
			\label{fig:weakMWexpc}
			}
   		\subfigure[Ours without loop closure]{
			\includegraphics[width=0.46\columnwidth]{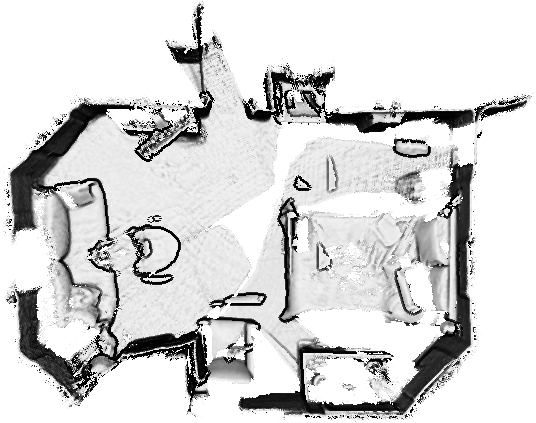}
			\label{fig:weakMWexpd}
			}\\
	\end{center} 
	\caption{Reconstruction results in the weak Manhattan world scene.} \vspace{-12pt}\label{fig:weakMWexp}
\end{figure}
%
\vspace{1.0mm}\noindent\textbf{Qualitative evaluation:} We compare reconstructed results of some selected methods~\cite{Whelan:2012,Xiao:2013,Choi:2015} in challenging real-world datasets provided by Xiao~\etal~\cite{Xiao:2013}. As shown in Fig.~\ref{fig:realdata}, Kintinuous, SUN3D SFM, and the Choi \etal method cannot preserve the genuine structure of walls in the real-world datasets. 
These results of the methods except our method show curved structures as well as largely distorted walls. Moreover, the SUN3D SFM shows noisy 3D points along the wall region. 
However, in the presence of a large amount of errors, the proposed method shows significantly improved results as shown in Fig.~\ref{fig:iterativeopt_e} and \ref{fig:iterativeopt_j}. The effect of the layout-constrained registration can be found clearly in real-world datasets.

For the qualitative evaluation using a weak Manhattan world scene, the results of \textit{hotel\_stb\_scan3} dataset are used for comparison with other methods.
As shown in Fig.~\ref{fig:weakMWexpa} and \ref{fig:weakMWexpb}, the Choi \etal method and ours yield good reconstruction results in the weak Manhattan world scene. 
In addition, we performed the experiments only using the first 1,000 frames out of 3,756 frames of the same dataset so that loop closures were not detected. 
Here, it is observed that the result reconstructed by the Choi \etal method (Fig.~\ref{fig:weakMWexpc}) is slightly bent, whereas the result by our method (Fig.~\ref{fig:weakMWexpd}) is similar to the result by applying loop closing (Fig.~\ref{fig:weakMWexpb}), with the aid of layout information.
Similarly, loop closures are not detected properly in Fig.~\ref{fig:iterativeopt_d} and \ref{fig:iterativeopt_i}, and therefore, reconstruction results are poor.
Nevertheless, the proposed method recovers rectangular shapes of the scene in Fig.~\ref{fig:iterativeopt_e} and \ref{fig:iterativeopt_j}. 
Slight quality degradation can occasionally happen if the reconstructed model is overfitted, \textit{e.g.}, as in \textit{Office1}. 
However, the proposed method improves the reliability of reconstruction while retaining accurate reconstruction results. 

\vspace{1.0mm}\noindent\textbf{Computational complexity:} The proposed method was implemented mixedly in MATLAB and C++ and ran on 2.6 GHz CPU with single core. For the \textit{Livingroom1} dataset, the total computational time of our method was about 3 hours, which was two times of that of the Choi \etal method implemented in C++.
The largest burden in the process is the dominant plane extraction step. The process of the dominant plane extraction has a complexity of $O(nml)$ since it computes the distance between $n$ plane hypotheses and $m$ points in $l$ fragments.
Although the proposed method is an offline method as \cite{Xiao:2013,Choi:2015} and is slower than the state-of-the-art methods, we explicitly assure that our method offers more reliable, robust, and accurate indoor 3D reconstruction results in comparison with the real-time methods~\cite{Whelan:2012,Whelan:2015} and other state-of-the-art offline methods~\cite{Xiao:2013,Choi:2015}.

\section{Conclusion}

We have presented an indoor 3D reconstruction algorithm that alternately resolves  two complementary problems, scene layout estimation and global registration, in iterative fashion. Given initially registered scene fragments, we estimate the envelope of a scene through hierarchical clustering and energy-based multi-model fitting and find a minimum set of planes that best describe the entire scene. From these plane hypotheses, we extract the scene layout that surrounds the entire point cloud, assuming that they coincide with walls, a floor, and a ceiling. We exploit the scene layout information to obtain globally consistent reconstruction results by constraining the global registration problem with scene layout information. We verified the superiority of the proposed method by using various datasets, including a challenging real-world dataset.


\section*{Acknowledgement}

This work was supported by the National Research Foundation of Korea (NRF) grant (No. NRF-2015R1A2A1A01005455) and `The Cross-Ministry Giga KOREA Project' grant (GK17P0300, Real-time 4D reconstruction of dynamic objects for ultra-realistic service) funded by the Korea government(MSIT).

{\small
\bibliographystyle{ieee}
\bibliography{egbib}

\begin{thebibliography}{10}\itemsep=-1pt

\bibitem{Arrigoni:2016}
F.~Arrigoni, B.~Rossi, and A.~Fusiello.
\newblock Global registration of 3d point sets via lrs decomposition.
\newblock In {\em ECCV}, 2016.

\bibitem{Ataer:2013}
E.~Ataer-Cansizoglu, Y.~Taguchi, S.~Ramalingam, and T.~Garaas.
\newblock Tracking an rgb-d camera using points and planes.
\newblock In {\em ICCV Workshops}, 2013.

\bibitem{Bergevin:1996}
R.~Bergevin, M.~Soucy, H.~Gagnon, and D.~Laurendeau.
\newblock Towards a general multi-view registration technique.
\newblock {\em TPAMI}, 18(5):540--547, 1996.

\bibitem{Besl:1992}
P.~J. Besl and N.~D. McKay.
\newblock A method for registration of 3-d shapes.
\newblock {\em TPAMI}, 14(2):239--256, Feb. 1992.

\bibitem{Borrmann:2008}
D.~Borrmann, J.~Elseberg, K.~Lingemann, A.~Nüchter, and J.~Hertzberg.
\newblock Globally consistent 3d mapping with scan matching.
\newblock {\em Robotics and Autonomous Systems}, 56(2):130--142, 2008.

\bibitem{Boykov:2001}
Y.~Boykov, O.~Veksler, and R.~Zabih.
\newblock Fast approximate energy minimization via graph cuts.
\newblock {\em TPAMI}, 23(11):1222--1239, 2001.

\bibitem{Chen:2013}
J.~Chen, D.~Bautembach, and S.~Izadi.
\newblock Scalable real-time volumetric surface reconstruction.
\newblock {\em ACM Trans. Graph. (TOG)}, 32(4), July 2013.

\bibitem{Chen:1992}
Y.~Chen and G.~Medioni.
\newblock Object modeling by registration of multiple range images.
\newblock {\em Image and Vision Computing}, 10(3):145--155, 1992.

\bibitem{Choi:2015}
S.~Choi, Q.-Y. Zhou, and V.~Koltun.
\newblock Robust reconstruction of indoor scenes.
\newblock In {\em CVPR}, 2015.

\bibitem{Choi:2013}
W.~Choi, Y.~W. Chao, C.~Pantofaru, and S.~Savarese.
\newblock Understanding indoor scenes using 3d geometric phrases.
\newblock In {\em CVPR}, 2013.

\bibitem{Cupec:2015}
R.~Cupec, E.~K. Nyarko, D.~Filko, A.~Kitanov, and I.~Petrovi{\'c}.
\newblock Place recognition based on matching of planar surfaces and line
  segments.
\newblock {\em IJRR}, 34(4-5):674--704, 2015.

\bibitem{Curless:1996}
B.~Curless and M.~Levoy.
\newblock A volumetric method for building complex models from range images.
\newblock In {\em SIGGRAPH}, 1996.

\bibitem{Dou:2012}
M.~Dou, L.~Guan, J.-M. Frahm, and H.~Fuchs.
\newblock Exploring high-level plane primitives for indoor 3d reconstruction
  with a hand-held rgb-d camera.
\newblock In {\em ACCV}, 2012.

\bibitem{Fern:2013}
E.~Fern{\'a}ndez-Moral, W.~Mayol-Cuevas, V.~Ar{\'e}valo, and
  J.~Gonz{\'a}lez-Jim{\'e}nez.
\newblock Fast place recognition with plane-based maps.
\newblock In {\em ICRA}, 2013.

\bibitem{Furlan:2013}
A.~Furlan, D.~Miller, D.~G. Sorrenti, L.~Fei-Fei, and S.~Savarese.
\newblock Free your camera: 3d indoor scene understanding from arbitrary camera
  motion.
\newblock In {\em BMVC}, 2013.

\bibitem{Furukawa:2009}
Y.~Furukawa, B.~Curless, S.~Seitz, and R.~Szeliski.
\newblock Manhattan-world stereo.
\newblock In {\em CVPR}, 2009.

\bibitem{Geiger:2011}
A.~Geiger, C.~Wojek, and R.~Urtasun.
\newblock Joint 3d estimation of objects and scene layout.
\newblock In {\em NIPS}. 2011.

\bibitem{Gelfand:2005}
N.~Gelfand, N.~J. Mitra, L.~J. Guibas, and H.~Pottmann.
\newblock Robust global registration.
\newblock In {\em Symposium on geometry processing}, volume~2, 2005.

\bibitem{Lee:2010}
A.~Gupta, M.~Hebert, T.~Kanade, and D.~M. Blei.
\newblock Estimating spatial layout of rooms using volumetric reasoning about
  objects and surfaces.
\newblock In {\em NIPS}. 2010.

\bibitem{Henry:2013}
P.~Henry, D.~Fox, A.~Bhowmik, and R.~Mongia.
\newblock Patch volumes: Segmentation-based consistent mapping with rgb-d
  cameras.
\newblock In {\em International Conference on 3D Vision (3DV)}, 2013.

\bibitem{Heredia:2012}
F.~Heredia and R.~Favier.
\newblock Kinectfusion extensions to large scale environments.
\newblock \url{http://www.pointclouds.org/blog/srcs/fheredia/index.php}, 2012.
\newblock Online; accessed 11-May-2012.

\bibitem{Huber:2011}
P.~J. Huber.
\newblock {\em Robust statistics}.
\newblock Springer Berlin Heidelberg, 2011.

\bibitem{Iwanowski:2007}
M.~Iwanowski.
\newblock Morphological boundary pixel classification.
\newblock In {\em International Conference on "Computer as a Tool"}, 2007.

\bibitem{Liu:2014}
Y.~Liu, W.~Zhou, Z.~Yang, J.~Deng, and L.~Liu.
\newblock Globally consistent rigid registration.
\newblock {\em Graphical Models}, 76(5):542--553, 2014.

\bibitem{Ma:2016}
L.~Ma, C.~Kerl, J.~Stueckler, and D.~Cremers.
\newblock Cpa-slam: Consistent plane-model alignment for direct rgb-d slam.
\newblock In {\em ICRA}, 2016.

\bibitem{Magri:2014}
L.~Magri and A.~Fusiello.
\newblock T-linkage: A continuous relaxation of j-linkage for multi-model
  fitting.
\newblock In {\em CVPR}, June 2014.

\bibitem{Newcombe:2011}
R.~A. Newcombe, S.~Izadi, O.~Hilliges, D.~Molyneaux, D.~Kim, A.~J. Davison,
  P.~Kohli, J.~Shotton, S.~Hodges, and A.~Fitzgibbon.
\newblock Kinectfusion: Real-time dense surface mapping and tracking.
\newblock In {\em ISMAR}, 2011.

\bibitem{Nie:2013}
M.~Nie{\ss}ner, M.~Zollh{\"o}fer, S.~Izadi, and M.~Stamminger.
\newblock Real-time 3d reconstruction at scale using voxel hashing.
\newblock {\em ACM Trans. Graph. (TOG)}, 32(6):169, 2013.

\bibitem{Nishino:2002}
K.~Nishino and K.~Ikeuchi.
\newblock Robust simultaneous registration of multiple range images.
\newblock In {\em ACCV}. 2002.

\bibitem{Oesau:2016}
S.~Oesau, F.~Lafarge, and P.~Alliez.
\newblock Planar shape detection and regularization in tandem.
\newblock In {\em Computer Graphics Forum}, volume~35, pages 203--215, 2016.

\bibitem{Papon:2013}
J.~Papon, A.~Abramov, M.~Schoeler, and F.~Worgotter.
\newblock Voxel cloud connectivity segmentation-supervoxels for point clouds.
\newblock In {\em CVPR}, 2013.

\bibitem{Pulli:1999}
K.~Pulli.
\newblock Multiview registration for large data sets.
\newblock In {\em Second International Conference on 3-D Digital Imaging and
  Modeling}. 1999.

\bibitem{Roth:2012}
H.~Roth and M.~Vona.
\newblock Moving volume kinectfusion.
\newblock In {\em BMVC}, 2012.

\bibitem{Rusu:2009}
R.~B. Rusu, N.~Blodow, and M.~Beetz.
\newblock Fast point feature histograms (fpfh) for 3d registration.
\newblock In {\em ICRA}, 2009.

\bibitem{Rusu:2011}
R.~B. Rusu and S.~Cousins.
\newblock {3D is here: Point Cloud Library (PCL)}.
\newblock In {\em ICRA}, 2011.

\bibitem{Moreno:2014}
R.~F. Salas{-}Moreno, B.~Glocker, P.~H.~J. Kelly, and A.~J. Davison.
\newblock Dense planar slam.
\newblock In {\em ISMAR}, 2014.

\bibitem{Shiratori:2015}
T.~Shiratori, J.~Berclaz, M.~Harville, C.~Shah, T.~Li, Y.~Matsushita, and
  S.~Shiller.
\newblock Efficient large-scale point cloud registration using loop closures.
\newblock In {\em International Conference on 3D Vision (3DV)}. 2015.

\bibitem{Steinbrucker:2013}
F.~Steinbrucker, C.~Kerl, and D.~Cremers.
\newblock Large-scale multi-resolution surface reconstruction from rgb-d
  sequences.
\newblock In {\em ICCV}, 2013.

\bibitem{Taguchi:2013}
Y.~Taguchi, Y.-D. Jian, S.~Ramalingam, and C.~Feng.
\newblock Point-plane slam for hand-held 3d sensors.
\newblock In {\em ICRA}, 2013.

\bibitem{Verdie:2015}
Y.~Verdie, F.~Lafarge, and P.~Alliez.
\newblock Lod generation for urban scenes.
\newblock {\em ACM TOG}, 34(3):30, 2015.

\bibitem{Whelan:2012}
T.~Whelan, M.~Kaess, M.~Fallon, H.~Johannsson, J.~Leonard, and J.~McDonald.
\newblock Kintinuous: Spatially extended {K}inect{F}usion.
\newblock In {\em RSS Workshop on RGB-D: Advanced Reasoning with Depth
  Cameras}, 2012.

\bibitem{Whelan:2013}
T.~Whelan, M.~Kaess, J.~J. Leonard, and J.~McDonald.
\newblock Deformation-based loop closure for large scale dense rgb-d slam.
\newblock In {\em IROS}, 2013.

\bibitem{Whelan:2015}
T.~Whelan, S.~Leutenegger, R.~S. Moreno, B.~Glocker, and A.~Davison.
\newblock Elasticfusion: Dense slam without a pose graph.
\newblock In {\em RSS}, 2015.

\bibitem{Williams:2000}
J.~A. Williams and M.~Bennamoun.
\newblock Simultaneous registration of multiple point sets using orthonormal
  matrices.
\newblock In {\em IEEE International Conference on Acoustics, Speech, and
  Signal Processing (ICASSP)}, 2000.

\bibitem{Xiao:2012}
J.~Xiao and Y.~Furukawa.
\newblock Reconstructing the world's museums.
\newblock In {\em ECCV}, 2012.

\bibitem{Xiao:2013}
J.~Xiao, A.~Owens, and A.~Torralba.
\newblock {SUN3D}: A database of big spaces reconstructed using sfm and object
  labels.
\newblock In {\em ICCV}, 2013.

\bibitem{Tang:2015}
J.~F. Y.~Tang.
\newblock Hierarchical multiview rigid registration.
\newblock {\em Computer Graphics Forum}, 34(5):77--87, 2015.

\bibitem{Zhang:2015}
Y.~Zhang, W.~Xu, Y.~Tong, and K.~Zhou.
\newblock Online structure analysis for real-time indoor scene reconstruction.
\newblock {\em ACM Trans. Graph. (TOG)}, 34(5):159:1--159:13, Nov. 2015.

\bibitem{Zhou:2014}
Q.-Y. Zhou and V.~Koltun.
\newblock Simultaneous localization and calibration: Self-calibration of
  consumer depth cameras.
\newblock In {\em CVPR}, 2014.

\bibitem{Zhou:2012}
Q.-Y. Zhou and U.~Neumann.
\newblock 2.5d building modeling by discovering global regularities.
\newblock In {\em CVPR}, pages 326--333, 2012.

\end{thebibliography}
}

\end{document}